\documentclass[runningheads]{llncs}

\usepackage{eccv}

\usepackage{eccvabbrv}

\usepackage{graphicx}
\usepackage{booktabs}
\usepackage{multirow}
\usepackage{mathtools, cases}

\usepackage[accsupp]{axessibility}  % Improves PDF readability for those with disabilities.

\usepackage{hyperref}

\usepackage{orcidlink}

\begin{document}

% ---------------------------------------------------------------
% TODO REVIEW: Replace with your title
\title{VLM-HOI: Vision Language Models for Interpretable Human-Object Interaction Analysis} 

% TODO REVIEW: If the paper title is too long for the running head, you can set
% an abbreviated paper title here. If not, comment out.
\titlerunning{VLM-HOI: Vision Language Model for Human-Object Interaction}

% TODO FINAL: Replace with your author list. 
% Include the authors' OCRID for the camera-ready version, if at all possible.
\author{Donggoo Kang$^1$ \and Dasol Jeong$^1$ \and Hyunmin Lee$^2$ \and Sangwoo Park$^1$ \and Hasil Park$^1$ \and Sunkyu Kwon$^2$ \and Yeongjoon Kim$^2$ \and Joonki Paik$^{1,2,*}$}

% TODO FINAL: Replace with an abbreviated list of authors.
\authorrunning{Kang et al.}
% First names are abbreviated in the running head.
% If there are more than two authors, 'et al.' is used.

% TODO FINAL: Replace with your institution list.
\institute{Department of $^1$Image and $^2$Artificial Intelligence \\
Chung-Ang University, Seoul, Korea \\
\email{\{dgkang, dasolj, dl218218, swpark, hspark, kwonsk, yjkim, paikj\}@ipis.cau.ac.kr}}

\maketitle

\begin{abstract}
The Large Vision Language Model (VLM) has recently addressed remarkable progress in bridging two fundamental modalities.
VLM, trained by a sufficiently large dataset, exhibits a comprehensive understanding of both visual and linguistic to perform diverse tasks.
To distill this knowledge accurately, in this paper, we introduce a novel approach that explicitly utilizes VLM as an objective function form for the Human-Object Interaction (HOI) detection task (\textbf{VLM-HOI}).
Specifically, we propose a method that quantifies the similarity of the predicted HOI triplet using the Image-Text matching technique.
We represent HOI triplets linguistically to fully utilize the language comprehension of VLMs, which are more suitable than CLIP models due to their localization and object-centric nature.
This matching score is used as an objective for contrastive optimization. 
To our knowledge, this is the first utilization of VLM language abilities for HOI detection. Experiments demonstrate the effectiveness of our method, achieving state-of-the-art HOI detection accuracy on benchmarks. We believe integrating VLMs into HOI detection represents important progress towards more advanced and interpretable analysis of human-object interactions.
\end{abstract}

\section{Introduction}
\label{sec:intro}

\begin{figure}
    \centering
    \includegraphics[width=0.9\linewidth]{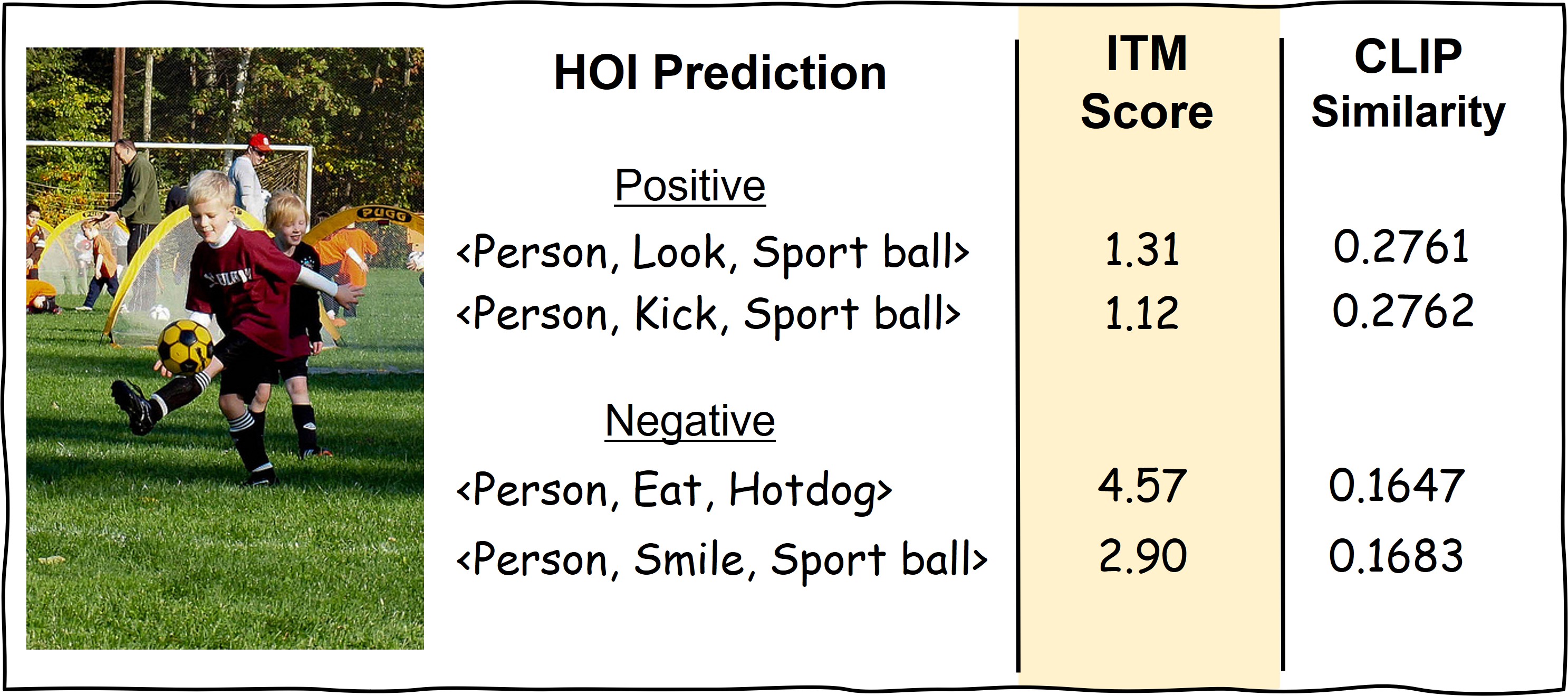}
    \caption{Comparison of Image-Text Matching scores and CLIP similarity for various HOI triplets.
    While CLIP struggles to capture semantic relationships due to its reliance on simple prompts, VLM models effectively distinguish between positive and negative HOI triplets despite not receiving complete sentences as input.}
    \label{fig:intro}
\end{figure}

Human-object interaction (HOI) detection plays a pivotal role in understanding the interactions between humans and objects within visual scenes. 
This task involves identifying and locating both the objects in an image or video and the specific interactions or actions that humans perform with those objects.
The ability to comprehend these intricate relationships is crucial for a wide range of applications, including image captioning\cite{yao2018exploring}, robotics\cite{xie2023visibility}, action recognition\cite{action_hoi}, scene understanding\cite{scene_hoi}.

HOI Detection involves two subtasks: 1) localizing the subject (human) and the target (object) of the interaction, and 2) classifying the interaction label that describes the relationship between them. 
For example, in an image of a person riding a bike, the subject is the person, the target is the bike, and the interaction label is ``ride''.
By recognizing the actions and interactions, such as ``riding a bicycle,'' ``cooking,'' or ``playing a musical instrument,'' HOI detection enables machines to have a more nuanced understanding of human behavior in visual data.
In this era of deep learning and the expanding availability of visual data, HOI detection has made remarkable progress\cite{CDN, fgahoi, gen_vlkt, hotr, muren, ppdm, qahoi, qpic, rlip, scg, STIP, upt, hoi_trans}. 
Deep neural networks have revolutionized the field, enabling the development of more accurate and robust models for recognizing human-object interactions. 
Additionally, the increasing availability of large-scale datasets and improved computational resources have further accelerated advances in this area.

In recent years, the field of artificial intelligence has witnessed remarkable progress in the integration of two fundamental modalities: vision and language\cite{blip, blip2, clip, chen2020uniter, li2019visualbert, li2021align, su2019vl, wang2020learning, yu2022coca, zhou2020unified}.
One of the most notable advancements in this domain has been the development of Large Vision Language Models (VLM), also known as the vision foundation model.
The early success of the foundation model is mainly due to the pre-training strategy that trains massive text datasets using self-supervised learning objectives\cite{bert, GPT3, palm}.
This means that the model is trained to learn general patterns and features from the data without the need for explicit labels. 
This is a much more efficient approach than supervised learning, which requires large amounts of labeled data.
Through the success of the text-driven foundation model, the vision foundation models also have achieved state-of-the-art results on a wide range of vision tasks, including image classification, object detection, and image captioning.
These results show the vision foundation models that pre-trained on vast and diverse datasets, exhibit a deep comprehension of both visual and linguistic information, enabling them to excel in a wide range of tasks that require the fusion of these modalities.

% \begin{figure}
%     \centering
%     \begin{subfigure}[b]{0.49\textwidth}
%          \centering
%          \includegraphics[width=\textwidth]{figures/gen_vlkt.jpg}
%          \caption{CLIP based approach\cite{gen_vlkt, hoiclip}}
%          \label{fig:y equals x}
%     \end{subfigure}
%     \hfill
%     \begin{subfigure}[b]{0.49\textwidth}
%          \centering
%          \includegraphics[width=\textwidth]{figures/ours_simple.jpg}
%          \caption{VLM-HOI}
%          \label{fig:three sin x}
%     \end{subfigure}
%     \caption{This figure compares the architecture of the proposed method to existing CLIP-based approaches like GenVLKT and HOIClip\cite{gen_vlkt, hoiclip}. Both CLIP-based approaches utilize CLIP for knowledge distillation, focusing on interactions through cosine similarity and employing two-branch decoders for interaction classification. In contrast, our proposed method leverages the VLM in a contrastive manner for knowledge distillation, integrating it directly into the loss function.}
%     \label{fig:enter-label}
% \end{figure}

\begin{figure}
    \centering
    \includegraphics[width=\textwidth]{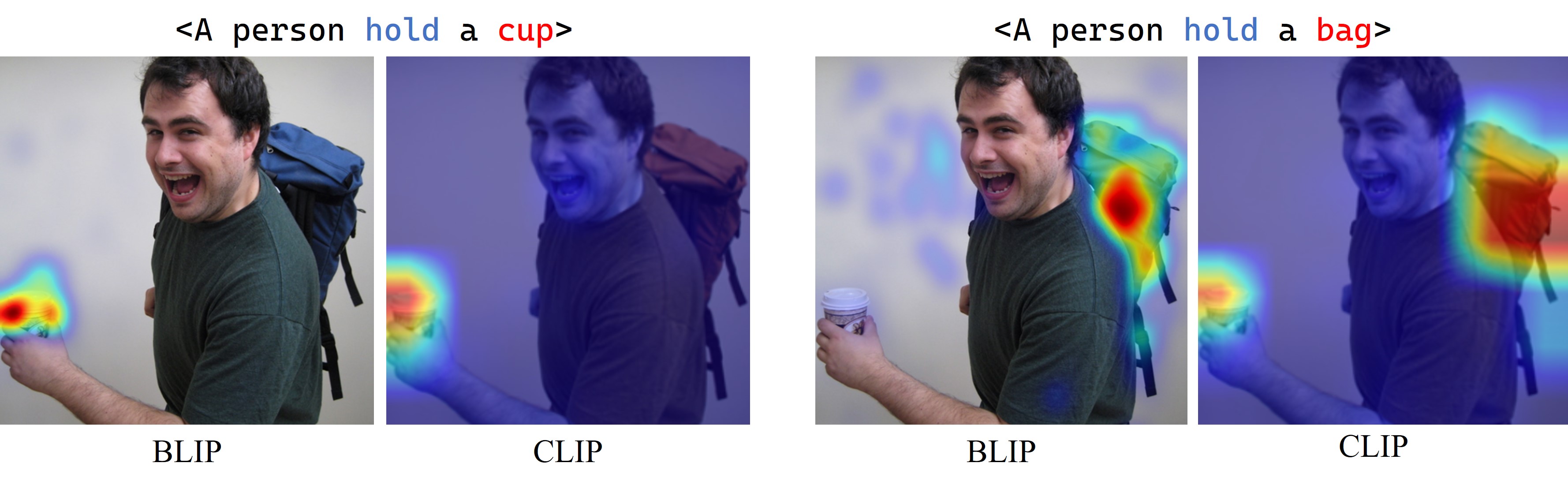}
    \caption{Visualization of attention maps for BLIP\cite{blip} and CLIP\cite{clip} models. Both models exhibit accurate localization in the first example. However, in the second example, CLIP focuses solely on individual word locations, failing to capture the broader context of the input sentence.}
    \label{fig:VLMvsCLIP}
\end{figure}

In light of this, we propose a novel approach that leverages the capabilities of large VLM as a distillation form for the HOI detection task (\textbf{VLM-HOI}). 
Our proposed method measures similarity in predicted HOI triplets, utilizing the concept of the Image-Text matching method. 
To achieve this, we represent HOI triplets in a linguistic form, capitalizing on the language understanding capabilities inherent in VLMs.
This linguistic representation allows us to harness the power of the VLM to comprehend and interpret the interactions between humans and objects in a more nuanced and context-aware manner.
Despite not being a complete sentence, VLM can compute the similarity between image and language, as shown in Figure.~\ref{fig:intro}.
To explicitly distill the knowledge of VLM to our approach, we utilize the ITM score computed from the Image-Text matching as our objective function, employing a contrastive learning framework.
This framework enables the network to understand HOI in arbitrary text form.
In our pursuit of advancing HOI detection, we conduct extensive experiments to evaluate the effectiveness of our proposed method. 

Experimental results reveal that our approach outperforms existing methods in terms of accuracy and robustness.
Through this work, we aim to contribute to the growing body of research at the intersection of vision and language, demonstrating the potential of VLMs as powerful tools for enhancing the understanding of complex visual scenarios involving human-object interactions.

The main contributions of this paper are summarized as follows:

\begin{itemize}
    \item We propose a novel approach that leverages the capabilities of large VLMs as a distillation form for the HOI detection task.
    \item We present a method that measures similarity in predicted HOI triplets, utilizing the concept of the Image-Text matching method.
    \item We develop a contrastive learning framework that enables the network to understand HOI in text form.
    \item We evaluate our proposed approach on two challenging HOI detection benchmarks and achieve state-of-the-art results.
\end{itemize}

\section{Related Work}
\label{sec:related_work}
\subsection{Human-Object Interaction Detection}

Human-object interaction (HOI) detection is an active area of research in computer vision. 
As noted in the introduction, HOI detection is a high-level task built on top of object detection. 
Existing HOI detection methods can be categorized into two main approaches: one-stage models and two-stage models.

One-stage models\cite{fgahoi, qpic} aim to detect human-object interactions in a single feedforward pass through a neural network.
These models take an image as input and directly output bounding boxes for humans, objects, and their interactions.
While computationally efficient, one-stage models struggle to model complex interactions and scale to large numbers of interaction categories.
In contrast, two-stage models\cite{qahoi, ppdm, CDN, scg, upt, gao2018ican} separate the tasks of detecting humans and objects from modeling their interactions. 
In the first stage, off-the-shelf object detectors are used to localize candidate humans and objects in the image.
The second stage then classifies their relationship based on appearance and spatial cues. 
By decomposing the problem, two-stage models are able to achieve higher accuracy but at the cost of reduced speed.

Another content \cite{rlip, gen_vlkt, hotr, muren, park2023viplo, zhang2023exploring} recent work has focused on improving both branches of HOI detection. 
For one-stage models, newer architectures incorporate attention mechanisms and graphical networks to better model interactions\cite{hotr, muren, park2023viplo}. 
For two-stage models, progress has been made in embedding space designs and developing robust spatial models\cite{zhang2023exploring}.

From another perspective, GEN-VLKT\cite{gen_vlkt} proposes a Visual-Linguistic Knowledge Transfer (VLKT) strategy that leverages CLIP to enhance interaction understanding. 
It uses CLIP text embeddings to initialize the HOI classifiers and mimics CLIP image features.
RLIP\cite{rlip, yuan2023rlipv2} proposes a pre-training strategy to train a robust backbone network by aligning the image representations of entities and relations with their corresponding text descriptions.
Both works are impressive approaches in that leverage VLM and pre-training strategy.
% Nonetheless, HOI detection remains a challenging problem, especially poor performance in rare classes in training datasets.

\subsection{Vision-Lauguage Model}

Integrating vision and language has been a long-standing goal in artificial intelligence. 
Early work focused on image captioning\cite{vinyals2015show}, generating textual descriptions of image contents. 
Recent years have seen rapid progress in visual question answering\cite{antol2015vqa, anderson2018bottom}, enabled by large-scale datasets\cite{chen2015microsoft} and deep neural encoder-decoder models\cite{xu2015show}.

More advanced vision-language tasks require a tighter integration between the visual and linguistic modalities. 
This has led to a surge of interest in unified multimodal representation models that can process both images and text within a single framework\cite{baltruvsaitis2019multimodal}.

One line of work explores joint embedding models to learn aligned vector representations for image regions and language fragments\cite{kiros2014unifying, karpathy2015deep}. However, these approaches do not explicitly model interactions between modalities.

Recently, pretrained language models, such as BERT\cite{bert} and GPT\cite{GPT3}, have significantly influenced the vision-language domain. 
Researchers have extended these models to handle multimodal inputs, leveraging their contextual understanding of text.
More recent methods like VisualBERT\cite{li2019visualbert} and LXMERT\cite{tan2019lxmert} utilize pretrained BERT weights for joint image-text comprehension.

Unified architectures like ViLBERT\cite{lu201912}, VL-BERT\cite{su2019vl}, UNITER\cite{chen2020uniter}, BLIP\cite{blip} integrate masked language modeling objectives alongside paired image-text prediction tasks within a single Transformer model.
These models set new state-of-the-art results on downstream tasks like visual question answering (VQA), visual reasoning, and image retrieval.

Specifically, we utilize BLIP as our vision-language model backbone. 
BLIP uses a flexible multimodal encoder-decoder model that can handle both understanding and generation tasks.
It also improves training data quality through generating new image captions and filtering noise. 
These advantages lead to strong performance.

Instruction-based VLMs like GPT-4V\cite{openai2023gpt4}, LLaVA\cite{liu2023llava}, and InstructBLIP\cite{instructblip} are a new type of large vision-language model that are trained to follow natural language instructions and prompts to perform various tasks.
A key advantage these models have over traditional fine-tuning is they can adapt to new tasks without needing gradient updates or lots of specific training data. The instruction format allows rapid adaptation.
However, still challenges exist around instruction following, ambiguity, high resource usage, and potential biases.

\section{Proposed Method}

\begin{figure}[t]
    \centering
    \includegraphics[width=\linewidth]{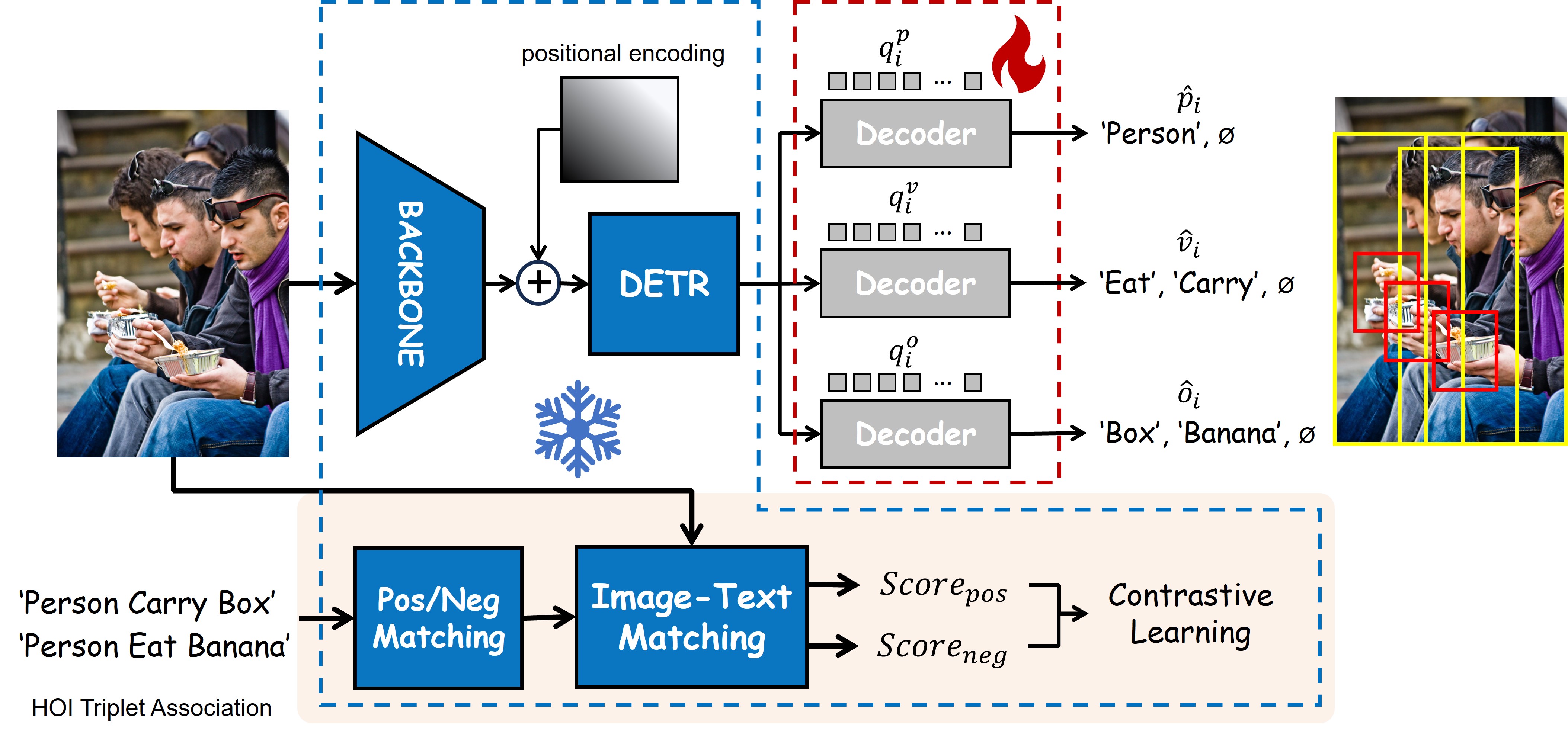}
    \caption{The overview of our proposed VLM-HOI. The network consists of a DETR-based encoder and a query-based transformer decoder. 
    Predicted HOI triplets are matched positive and negative 
    Then these sets are converted into text form. 
    The image-text matching task of VLM computes the matching score of these text sets.
    }
    \label{fig:overview}
\end{figure}

In this section, we introduce our proposed VLM-HOI method, which is illustrated in Figure~\ref{fig:overview}.
As shown, we adopt a detection transformer (DETR) as the feature encoder and object detector, following recent works \cite{hotr, muren, hoi_trans, qpic, upt, CDN, STIP}.
A query-based transformer decoder is also utilized to predict HOI triplets, as done in recent studies\cite{hotr, muren}.
The goal of this work is to effectively transfer the comprehensive language understanding capabilities of vision-language models (VLMs) to HOI detection.

In the following section, we first briefly define the baseline VLM model.
We then introduce a brief problem definition and present HOI triplet association technique that converts predicted HOI triplets into positive and negative text forms as inputs to the VLM.
These text sets are used to compute image-text matching scores with the VLM.
Subsequently, we introduce a contrastive learning loss to learn representations of VLM using these matching scores.

\subsection{Baseline VLM Model}

While large models like CLIP\cite{clip} excel at overall image-text similarity, their limitations in pinpointing specific objects become apparent. 
This paper takes a step forward, focusing on BLIP\cite{blip, blip2}, a VLM specifically designed for object-level understanding, and its potential as a powerful teacher model for knowledge distillation in localization tasks.
Prior works\cite{gen_vlkt, hoiclip} have utilized CLIP as a teacher model, achieving notable results.
However, we argue that BLIP offers distinct advantages for localization due to its inherent strengths.

\begin{itemize}
    \item \textbf{Object-centric Design:} BLIP learns representations for both images and object-specific descriptions, enabling pinpoint localization beyond CLIP's image-level approach.
    \item \textbf{Fine-grained Matching:} Unlike CLIP's simple cosine distance, BLIP employs sub-word level matching, connecting individual textual words to image regions, and facilitating precise object identification.
    \item \textbf{Specialized Training Data:} BLIP is pre-trained on object-centric datasets with detailed descriptions, equipping it with nuanced understanding of object features and spatial relationships, crucial for localization.
\end{itemize}

Figure~\ref{fig:intro}, suggests that the VLM model is able to better capture the semantic relationships between images and text for natural interactions. 
In contrast, the CLIP similarity scores are not as discriminative between positive and negative triplets. 
This suggests that CLIP may not be as effective at capturing the more nuanced relationships between images and text.

\subsection{Problem Definition}

Human-object interaction (HOI) detection aims to locate human and object instance in an image and classify their interaction relationship.
Formally, given an image $I$, the objective is to detect a set of $N$ interacted human-object pairs $\{(h_i, o_i, v_i)\}$ where $h_i \text{ and } o_i$ represent detected human and object.
$v_i$ denotes the interaction between $h_i \text{ and } o_i$. 

Each $h_i$ and $o_i$ is represented by a bounding box $b_i^h$ and $b_i^o$ respectively. 
The overall HOI prediction for the image can be denoted as:
\begin{equation}
    H=\{(b_i^h, b_i^o, o_i, v_i) | i\in 1,2,...,N\}
    \label{eq:problem_definition}
\end{equation}
where $H$ is the final output containing $N$ detected HOI triplets.
The HOI task can be decomposed into two sub-problems: Human and object detection $H_{det}=\{(b_i^h,b_i^o, o_i)\}$ and interaction classification given information of human and object, $H_{hoi}=\{(v_i|b_i^h,b_i^o, o_i)\}$.

\subsection{HOI Triplet Association}

The objective of the HOI triplet association is to convert these triplets into grounded natural language representations to leverage the language comprehension capabilities of vision-language models (VLMs).
Our model first predicts a set of $N$ predicted HOI triplets $T=\{(\hat{h}_1, \hat{o}_1, \hat{v}_1), ..., (\hat{h}_N, \hat{o}_N, \hat{v}_N)\}$ from the input image $I$, where $\hat{h}, \hat{o}, \hat{v}$ are the detected human, object and predicted interaction verb, respectively.

Given these predictions, we extract the detected object and interaction class names $o_i$ and $v_i$ from the classifier outputs for each triplet.
We exclude any triplets predicted as ``no interaction'' or ``no object'', denoted by $\oslash$, as determined by the Hungarian matching. 
For simplicity, we set all predicted human classes $\hat{h}$ as ``A person''. 
We then construct a positive sentence $S^{pos}_i$ for each triplet using the template:
% $S^{pos}_i = \text{"A person " + v_i + " a " + o_i} \quad \forall i \in [1,N]$
% \begin{numcases}{}
%     s^{+}_i = ``[\hat{h}_i]\,\, [\hat{o}_i]\,\, [\hat{v_i}]", & if ($\hat{h}_i, \hat{o}_i,\hat{v_i}$) $\in H$ \\
%     s^{-}_i = ``[\hat{h}_i]\,\, [\hat{o}_i]\,\, [\hat{v_i}]", & if ($\hat{h}_i, \hat{o}_i,\hat{v_i}$) $\notin H$
% \end{numcases}

\begin{equation}
\begin{split}
    s^{+}_i = [\text{``A person''} + \hat{v}_i + \text{`` a ''} + \hat{o_i}], \quad \text{if  } T \in H \\
    s^{-}_i = [\text{``A person''} + \hat{v}_i + \text{`` a ''} + \hat{o_i}], \quad \text{if  } T \notin H
\end{split}
\end{equation}

The examples of these sentences are described in Figure~\ref{fig:intro}.
The generated sentences may not be grammatically correct, since the class names in most HOI datasets are not designed to produce fluent phrases. Simply concatenating the human, verb, and object classes can result in unnatural or awkward language. However, these grounded sentences still provide useful context and supervision for knowledge distillation from the pre-trained VLM. The key elements of human, action, and object capture the core semantics of the visual HOI detections in a textual representation. While not linguistically flawless, this allows transferring relevant knowledge about humans, objects, and their interactions from the VLM to the HOI model.

The paired positive and negative sentences provide contrasting language context for the VLM to comprehend the visual HOI concept deeply.
This gives us a set of positive sentences $S^{pos} = \{s^{+}_1, ..., s^{+}_n\}$ and a set of negative sentences $S^{neg} = \{s^{-}_1, ..., s^{-}_m\}$ that are semantically grounded in the visual HOI predictions. 
By optimizing the VLM representations on these contrasting text sets, we enable the model to learn fine-grained HOI concepts based on its language priors. 
This facilitates transferring the VLM's comprehensive language understanding to improve HOI prediction.

\subsection{Image-Text Matching based Knowledge Distillation}

% \begin{figure}
%     \centering
%     \includegraphics[width=0.5\linewidth]{figures/contrastive.jpg}
%     \caption{The proposed image-text contrastive loss helps the HOI network to learn more efficiently, especially for rare classes. This is because the pre-trained VLM has a well-defined distribution over both image and language domains, which can guide the HOI network to find the optimal distribution for its own parameters.}
%     \label{fig:contrastive_learning}
% \end{figure}

% \begin{figure}[tb]
% \centering
%   \begin{subfigure}{0.4\textwidth}
%     \includegraphics[width=\textwidth]{figures/contrastive_orig.jpg}
%     \caption{Optimization progress of Baseline}
%     \label{fig:short-a}
%   \end{subfigure}
%   \begin{subfigure}{0.4\textwidth}
%     \includegraphics[width=\textwidth]{figures/contrastive_ours.jpg}
%     \caption{Another example of a subfigure}
%     \label{fig:short-b}
%   \end{subfigure}
%   \caption{The proposed image-text contrastive loss helps the HOI network to learn more efficiently, especially for rare classes. This is because the pre-trained VLM has a well-defined distribution over both image and language domains, which can guide the HOI network to find the optimal distribution for its own parameters.}
%   \label{fig:contrastive_learning}
% \end{figure}

To harness the image-text matching capabilities of VLMs, we compute similarity scores between a given image $I$ and paired sentence sets: $S^{\text{pos}}$ for positive (correct) associations and $S^{\text{neg}}$ for negative (incorrect) ones. 
The calculation is formalized as:

\begin{equation}
    \text{sim}(I, S) = \text{VLM}_{\text{itm}}(I, S),
    \label{eq:similarity}
\end{equation}
where $\text{sim}(I, S)$ denotes the image-text similarity score obtained from the VLM for the image $I$ and sentence set $S$.

Ideally, the similarity scores of positive sets should be as close to zero as possible, while the similarity scores of negative sets should be as high as possible.
However, Figure~\ref{fig:intro} illustrates that similarity scores cannot be zero for positive triplets.
This ensures that the loss is never zero, even when all predictions are correct.
We will discuss later how this problem inhibits the optimization of other losses.

To explicitly distill the rich knowledge encoded in the VLM's parameters, we employ these scores in a contrastive learning loss:
\begin{equation}
    \mathcal{L}_{\text{ITM}} =\sum^n_{i=0}(\max(0,\alpha - \text{sim}(I, s^+_i))) + \sum^m_{j=0}(\text{sim}(I, s^-_j))
\label{eq:itm_los}
\end{equation}
Here, $\alpha$ serves as positive margins that anchor the contrastive loss. 
This loss function is structured to optimize the pairing of images with their corresponding positive sentences while disassociating them from negative ones.

The key insight is that the pre-trained vision-language model has learned a robust joint distribution over both visual and textual modalities through extensive pre-training. 
By optimizing our HOI network to align with the VLM's image-text similarities via contrastive loss, we essentially distill this strong prior knowledge into our model. 
This guides the HOI network to find an optimal parameter configuration that encodes visual-linguistic concepts effectively. 
This distillation leads to more efficient optimization and substantial gains for rare HOI categories that lack sufficient training examples. 
In essence, the VLM's contextual knowledge helps regularize the HOI model, reducing overfitting and improving generalization.

\subsection{Training and Inference}

Following prior query-based HOI detection methods\cite{hotr, muren, CDN}, we utilize the Hungarian matching algorithm to associate predicted triplets with ground truth triplets.
As illustrated in Figure~\ref{fig:overview}, our network consists of a DETR-based encoder and three parallel decoder branches with task-specific queries $Q^p$, $Q^a$, and $Q^o$ for human, object, and interaction prediction respectively.
We select BLIP\cite{blip} as the VLM to compute image-text matching scores because it has higher performance relative to other methods in terms of computational efficiency.
In principle, any sufficiently large VLM can be utilized and continue to improve with more compute.

In addition to the contrastive image-text matching loss $\mathcal{L}_{\text{ITM}}$, we use a standard HOI detection loss $\mathcal{L}_{\text{HOI}}$ defined as:
\begin{equation}
    \mathcal{L_\text{HOI}} = \lambda_1\mathcal{L_\text{L1}} + \lambda_2\mathcal{L_\text{GIoU}} + \lambda_3\mathcal{L_\text{oc}} + \lambda_4\mathcal{L_\text{ic}}.
    \label{eq:hoi_loss}
\end{equation}
Where $\mathcal{L_\text{L1}}$ and $\mathcal{L_\text{GIoU}}$ are regression losses for predicting human and object bounding boxes, $\mathcal{L_\text{oc}}$ is a classification loss for detecting object categories, $\mathcal{L_\text{ic}}$ is a verb classification loss for interaction predictions, and $\lambda_i$ is weighting hyper-parameter.

We train our model end-to-end by jointly optimizing the HOI detection loss $\mathcal{L}_\text{{HOI}}$ and the image-text matching knowledge distillation loss $\mathcal{L}_{\text{ITM}}$:

\begin{equation}
    \mathcal{L}_{\text{total}} = \mathcal{L}_{\text{HOI}} + \mathcal{L}_{\text{ITM}}.
\end{equation}

The VLM is kept frozen during training to provide fixed similarity computations.
At inference time, we simply feed an image into our trained model to generate the detected HOI triplets. 
The VLM and image-text matching loss are only used during training for knowledge transfer and are not needed at test time.
Therefore, the number of learnable parameters are same as the baseline method.

\section{Experimental Results}

\subsection{Implementation Details}

Our model uses a ResNet-50 CNN backbone followed by a 6-layer transformer encoder, with $L=6$ parallel prediction branches. 
We set $N=64$ queries for HICO-DET and $N=100$ for V-COCO following prior work\cite{CDN, muren}. 
The loss weights $\lambda_i$ are set to 2.5, 1, 1, 1 for $\mathcal{L}_{\text{L1}}$, $\mathcal{L}_{\text{GIoU}}$, $\mathcal{L}_{\text{oc}}$, and $\mathcal{L}_{\text{ic}}$ respectively. 
We initialize from a DETR model\cite{detr} pretrained on COCO\cite{coco} and optimize using AdamW\cite{adamw} with weight decay 1e-4. 
The CNN backbone has a learning rate of 1e-5 while all other components use 1e-4, trained for 100 epochs. 
For V-COCO, the CNN weights are frozen to prevent overfitting and the learning rate is reduced to 4e-5. 
All experiments use a batch size of 4 on 4 RTX 3090ti GPUs. 
These optimized hyperparameters enable efficient end-to-end training of our method.

\subsection{Dataset}

We conduct experiments on two standard HOI detection benchmarks: HICO-DET\cite{hico} and V-COCO\cite{vcoco}. 

HICO-DET consists of 47,776 images, with 38,118 for training and 9,658 for testing. 
It contains 600 HOI categories composed of 80 object classes and 117 action verbs. 

V-COCO is a subset of 10,396 COCO images, split into 5,400 train and 4,964 test images. 
It has 29 action classes including 4 body motions without object interactions, and the same 80 objects as HICO-DET. 
In total, V-COCO has 263 unique HOI triplets.

\subsection{Evaluation Metric}

Following the evaluation protocol from\cite{CDN, muren}, we use mean Average Precision (mAP) as the metric for measuring HOI detection performance. 
A predicted triplet is considered a true positive if: 1) the detected human and object boxes have over 50\% IOU with ground truth, and 2) the predicted interaction categories match the labels. 

On HICO-DET, we report mAP on the full 600 classes, 138 rare classes, and 462 non-rare classes. 
For V-COCO, we evaluate on S1 with 29 actions including body motions, and S2 with 25 actions excluding no-object categories. 
By benchmarking on these diverse splits, we comprehensively analyze our method's HOI detection capabilities.

\subsection{Comparison with State-of-the-Art}

\begin{table*}[htb!]
    \centering
    \begin{tabular}{lcccccc}
    \toprule
    \toprule
         \multirow{2}{*}{\text{\quad Method \quad}} & \multicolumn{3}{c}{Default} & \multicolumn{3}{c}{Known Object} \\
         \cmidrule(lr){2-4} \cmidrule(lr){5-7}
         & \text{\quad Full \quad} & \text{\quad Rare \quad} & Non-Rare & \text{\quad Full \quad} & \text{\quad Rare \quad} & Non-Rare \\
         \midrule
         IDN\cite{IDN} & 24.58 & 20.33 & 25.86 & 27.89 & 23.64 & 29.16 \\       
         \cmidrule(lr){1-1} \cmidrule(lr){2-2} \cmidrule(lr){3-3} \cmidrule(lr){4-4} \cmidrule(lr){5-5} \cmidrule(lr){6-6} \cmidrule(lr){7-7}
         HOTR\cite{hotr} & 25.10 & 17.34 & 27.42 & - & - & - \\
         \cmidrule(lr){1-1} \cmidrule(lr){2-2} \cmidrule(lr){3-3} \cmidrule(lr){4-4} \cmidrule(lr){5-5} \cmidrule(lr){6-6} \cmidrule(lr){7-7}
         HOI-Trans\cite{hoi_trans}& 26.61 & 19.15 & 28.84 & 29.13 & 20.98 & 31.57 \\
         \cmidrule(lr){1-1} \cmidrule(lr){2-2} \cmidrule(lr){3-3} \cmidrule(lr){4-4} \cmidrule(lr){5-5} \cmidrule(lr){6-6} \cmidrule(lr){7-7}
         QPIC\cite{qpic} & 29.07 & 21.85 & 31.23 & 31.68 & 24.14 & 33.93 \\
         \cmidrule(lr){1-1} \cmidrule(lr){2-2} \cmidrule(lr){3-3} \cmidrule(lr){4-4} \cmidrule(lr){5-5} \cmidrule(lr){6-6} \cmidrule(lr){7-7}
         MSTR\cite{mstr} & 31.17 & 25.31 & 32.92 & 34.02 & 28.83 & 35.57 \\
         \cmidrule(lr){1-1} \cmidrule(lr){2-2} \cmidrule(lr){3-3} \cmidrule(lr){4-4} \cmidrule(lr){5-5} \cmidrule(lr){6-6} \cmidrule(lr){7-7}
         CDN\cite{CDN} & 32.07 & 27.19 & 33.53 & 34.79 & 29.48 & 36.38 \\
         \cmidrule(lr){1-1} \cmidrule(lr){2-2} \cmidrule(lr){3-3} \cmidrule(lr){4-4} \cmidrule(lr){5-5} \cmidrule(lr){6-6} \cmidrule(lr){7-7} 
         STIP\cite{STIP} & 32.22 & 28.15 & 33.43 & 35.29 & 31.43 & 36.45 \\
         \cmidrule(lr){1-1} \cmidrule(lr){2-2} \cmidrule(lr){3-3} \cmidrule(lr){4-4} \cmidrule(lr){5-5} \cmidrule(lr){6-6} \cmidrule(lr){7-7}
         UPT\cite{upt}& 32.62 & 28.62 & 33.81 & 36.08 & 31.41 & 37.47 \\
         \cmidrule(lr){1-1} \cmidrule(lr){2-2} \cmidrule(lr){3-3} \cmidrule(lr){4-4} \cmidrule(lr){5-5} \cmidrule(lr){6-6} \cmidrule(lr){7-7}
         MUREN\cite{muren} & 32.87 & 28.67 & 34.12 & 35.52 & 30.88 & 36.91 \\
         \cmidrule(lr){1-1} \cmidrule(lr){2-2} \cmidrule(lr){3-3} \cmidrule(lr){4-4} \cmidrule(lr){5-5} \cmidrule(lr){6-6} \cmidrule(lr){7-7}
         GEN-VLKT\cite{gen_vlkt} & 33.75 & 29.25 & 35.10 & 36.78 & 32.75 & \textbf{37.99} \\
         \midrule
         \textbf{VLM-HOI} & \textbf{34.25} & \textbf{30.22} & \textbf{35.20} & \textbf{36.88} & \textbf{33.30} & 37.75 \\
    \bottomrule
    \bottomrule
    \end{tabular}
    \caption{Comparison evaluation of our approach against state-of-the-art methods on the HICO-DET benchmark, excluding comparisons to other Swin Transformer-based models for fair analysis with the same backbone.}
    \label{tab:hico}
\end{table*}

We extensively evaluate our VLM-HOI method against state-of-the-art approaches on two standard HOI detection benchmarks - HICO-DET\cite{hico} and V-COCO\cite{vcoco}.

On HICO-DET (Table~\ref{tab:hico}), our proposed knowledge distillation method improves upon GEN-VLKT\cite{gen_vlkt}, which also utilizes external knowledge but only achieves 33.75\% and 36.78\% mAP on default and known object settings. In contrast, our VLM-HOI obtains 33.64\% and 36.88\% mAP, highlighting the benefits of our image-text matching objective.

Analyzing the long-tail distribution, the proposed VLM-HOI demonstrates consistent gains over GEN-VLKT, with improvements of 0.62\% and 0.55\% mAP on rare interactions. This validates our approach of transforming visual detections into grounded text for optimized knowledge transfer from VLMs.

Similarly, on V-COCO (Table~\ref{tab:v-coco}), the proposed method obtains 69.5\% and 72.1\% mAP on Scenario 1 and 2, surpassing all previous methods.
Here, VLM-HOI outperforms GEN-VLKT by large margins of 7.1\% AP on Scenario 1 and 7.7\% AP on Scenario 2.
This highlights the benefits of our image-text matching objective for richer knowledge transfer.

Notably, we improve over MUREN\cite{muren}, which uses the same transformer backbone as our method. This shows that the performance gains come from our proposed VLM knowledge distillation, rather than just model architecture.

\begin{figure}[ht]
    \centering
    \includegraphics[width=0.9\textwidth]{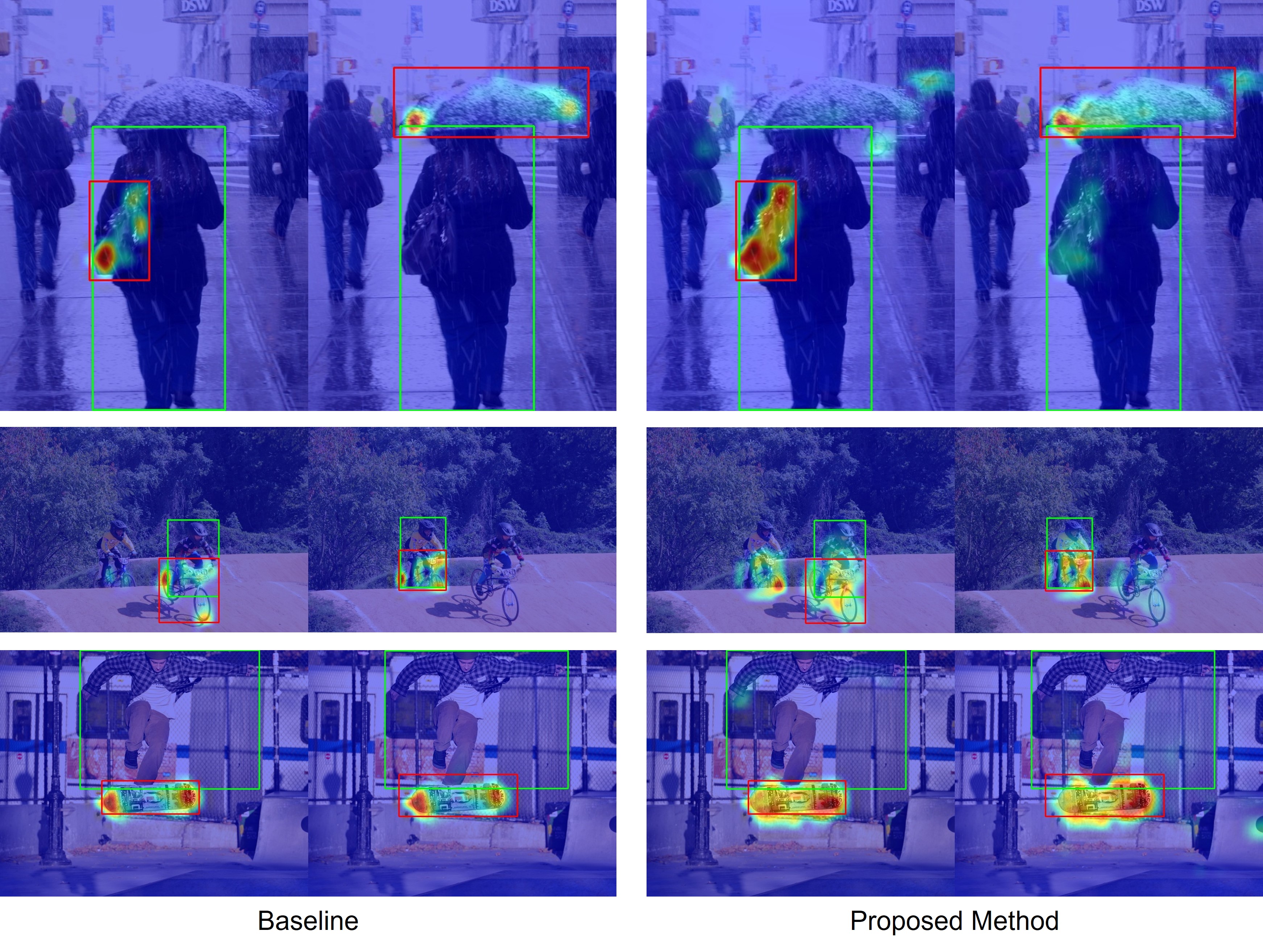}
    \caption{Comparison of Baseline\cite{muren} and proposed method with example queries. Given a verb query $q^v_i$, we visualize the top two most confident predictions, including their corresponding activation maps and bounding boxes.}
    \label{fig:enter-label}
\end{figure}

\begin{table}
\renewcommand{\arraystretch}{0.9}
\setlength{\tabcolsep}{17pt}
    \centering
    \begin{tabular}{lcc}
    \toprule
    \toprule
    \quad \quad Method & $\text{AP}^{\#1}_{\text{role}}$ & $\text{AP}^{\#2}_{\text{role}}$ \\
    \midrule
    GG-Net\cite{GG-Net} & 54.7  & - \\
    \cmidrule(lr){1-1} \cmidrule(lr){2-2} \cmidrule(lr){3-3}
    HOTR\cite{hotr} & 55.2  & 64.4 \\
    \cmidrule(lr){1-1} \cmidrule(lr){2-2} \cmidrule(lr){3-3}
    QPIC-R50\cite{qpic} & 58.8  & 61.0 \\
    \cmidrule(lr){1-1} \cmidrule(lr){2-2} \cmidrule(lr){3-3}
    GEN-VLKT\cite{gen_vlkt} & 62.4  & 64.4 \\
    \cmidrule(lr){1-1} \cmidrule(lr){2-2} \cmidrule(lr){3-3}
    CDN\cite{CDN} & 62.4  & 64.4 \\
    \cmidrule(lr){1-1} \cmidrule(lr){2-2} \cmidrule(lr){3-3}
    UPT\cite{upt} & 59.0  & 64.5 \\
    \cmidrule(lr){1-1} \cmidrule(lr){2-2} \cmidrule(lr){3-3}
    STIP\cite{STIP} & 66.0  & 70.7 \\
    \cmidrule(lr){1-1} \cmidrule(lr){2-2} \cmidrule(lr){3-3}
    DisTR\cite{DisTR} & 66.2  & 68.5 \\
    \cmidrule(lr){1-1} \cmidrule(lr){2-2} \cmidrule(lr){3-3}
    MUREN\cite{muren} & 66.5  & 68.7 \\
    \midrule
    \textbf{VLM-HOI} & \textbf{67.7} & \textbf{70.9} \\
    \bottomrule
    \bottomrule
    \end{tabular}
    \caption{Comparison evaluation of our approach against state-of-the-art methods on the V-COCO benchmark}
    \label{tab:v-coco}
\end{table}

\subsection{Ablation Study}
\textbf{Effects of the Margin}: A critical hyperparameter in our image-text contrastive loss is the positive sample margin $\alpha$. 
This margin controls the lower bound on the similarity scores between positive text prompts and the image. 
Intuitively, it determines how close we want to pull the positive grounded sentence representations to the image representation.

\begin{table}[h!]
\centering
\begin{minipage}{0.48\textwidth}
    \centering
    \setlength{\tabcolsep}{18pt}
    \begin{tabular}{lcc}
    \toprule
    $\alpha$ & $\text{AP}^{\#1}_{\text{role}}$ & $\text{AP}^{\#2}_{\text{role}}$ \\
    \midrule
    0  & 66.80 & 70.53 \\
    \cmidrule(lr){1-1} \cmidrule(lr){2-2} \cmidrule(lr){3-3}
    1  & \textbf{67.73} & \textbf{70.91} \\
    \cmidrule(lr){1-1} \cmidrule(lr){2-2} \cmidrule(lr){3-3}
    2  &  67.13 & 70.69 \\
    \bottomrule
    \end{tabular}
    \caption{Analysis of the positive margin hyperparameter $\alpha$ in the image-text contrastive loss on V-COCO.}
    \label{tab:alpha}
\end{minipage}\hfill
\begin{minipage}{0.48\textwidth}
    \centering
    \setlength{\tabcolsep}{16pt}
    \begin{tabular}{rcc}
    \toprule
    Prompt & $\text{AP}^{\#1}_{\text{role}}$ & $\text{AP}^{\#2}_{\text{role}}$ \\
    \midrule
    Verb  & 67.51 & 70.27 \\
    \cmidrule(lr){1-1} \cmidrule(lr){2-2} \cmidrule(lr){3-3}
    Object  & 67.29 & 70.52 \\
    \cmidrule(lr){1-1} \cmidrule(lr){2-2} \cmidrule(lr){3-3}
    Full  &  \textbf{67.73} & \textbf{70.91} \\
    \bottomrule
    \end{tabular}
    \caption{Performance comparison of different grounded prompt structures for converting predicted HOI triplets to text.}
    \label{tab:prompt_variation}
\end{minipage}
\end{table}

If $\alpha$ is too small or zero, the loss will not effectively separate positive and negative samples, as the lower bound on positive scores is too low. 
On the other hand, if $\alpha$ is too large, it can overly restrict and dominate the loss landscape.

Table~\ref{tab:alpha} shows the impact of $\alpha$ by evaluating models trained with different values. 
As shown in Figure~\ref{fig:intro}, the typical similarity scores between positive triplets and the image range from 1 to 2. Setting $\alpha=0$ gives a mAP drop of 1\% compared to $\alpha=1$, indicating that a zero margin fails to sufficiently separate distributions.

Increasing $\alpha = 2$ also decreases performance by 0.6\% mAP, suggesting that too large of a margin overly restricts the optimization. 
The best balance is achieved with $\alpha=1$, which aligns well with the expected positive sample score distribution.

The positive margin hyperparameter should be tuned to match the anticipated similarity score range, in order to enable effective optimization. 
This provides a principle for setting this important variable when applying our image-text contrastive loss to new domains beyond HOI detection.

\noindent\textbf{Effects of Prompt Variation}: In our main experiments, we utilize full grounded sentences to represent the predicted HOI triplets, e.g. ``A person hold a tennis racket''. 
Here we analyze the impact of using only partial prompts without the human/object classes.

As shown in Table~\ref{tab:prompt_variation}, using just the interaction verb as the prompt (``is holding'') decreases performance by 1.2\% mAP compared to the full sentence. This demonstrates that grounding the textual prompt with the detected human/object classes provides useful context and regularization for the VLM.

We also experiment with a prompt containing just the human and object (``A person tennis racket''). This achieves a 0.8\% lower mAP than the full prompt, showing that including the interaction verb is important for capturing the HOI semantics.

Overall, the full grounded sentence prompt leads to the best knowledge transfer from the VLM to our HOI detection model. 
The human, object, and verb provide complementary contextual cues that, when combined, enable optimized distillation of the VLM's visual-linguistic knowledge.

\begin{figure*}
    \centering
    \begin{subfigure}[b]{0.75\linewidth}
         \centering
         \includegraphics[width=\textwidth]{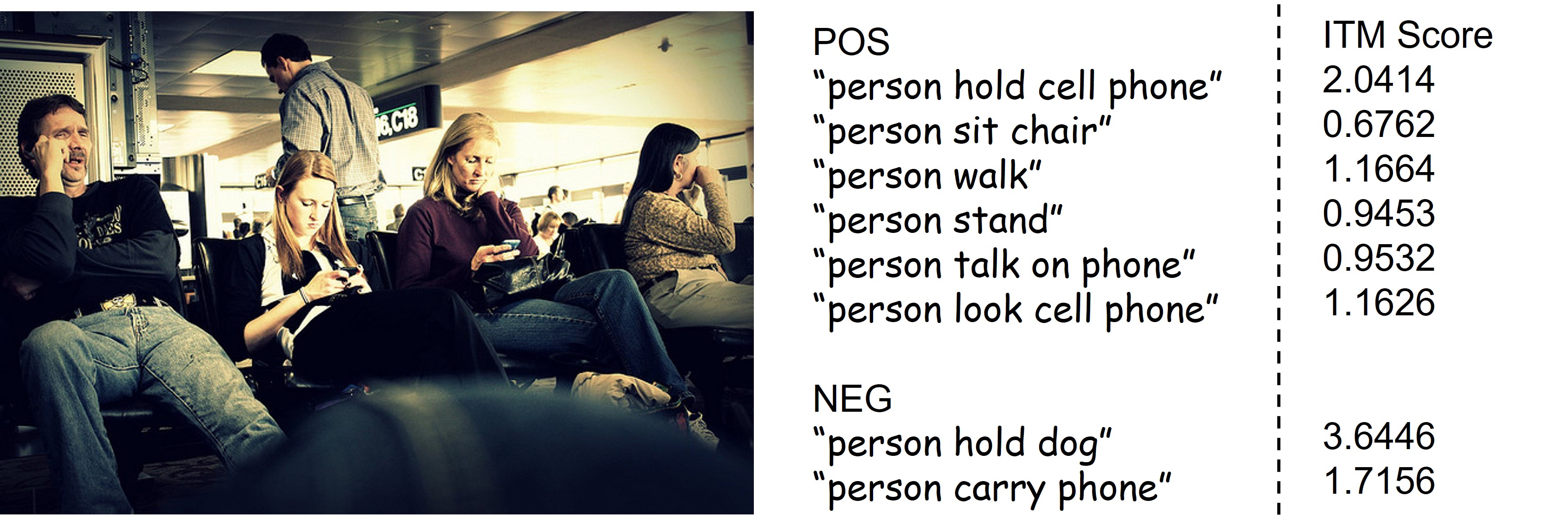}
         \label{fig:y equals x}
    \end{subfigure}
    \hfill
    \begin{subfigure}[b]{0.75\linewidth}
         \centering
         \includegraphics[width=\textwidth]{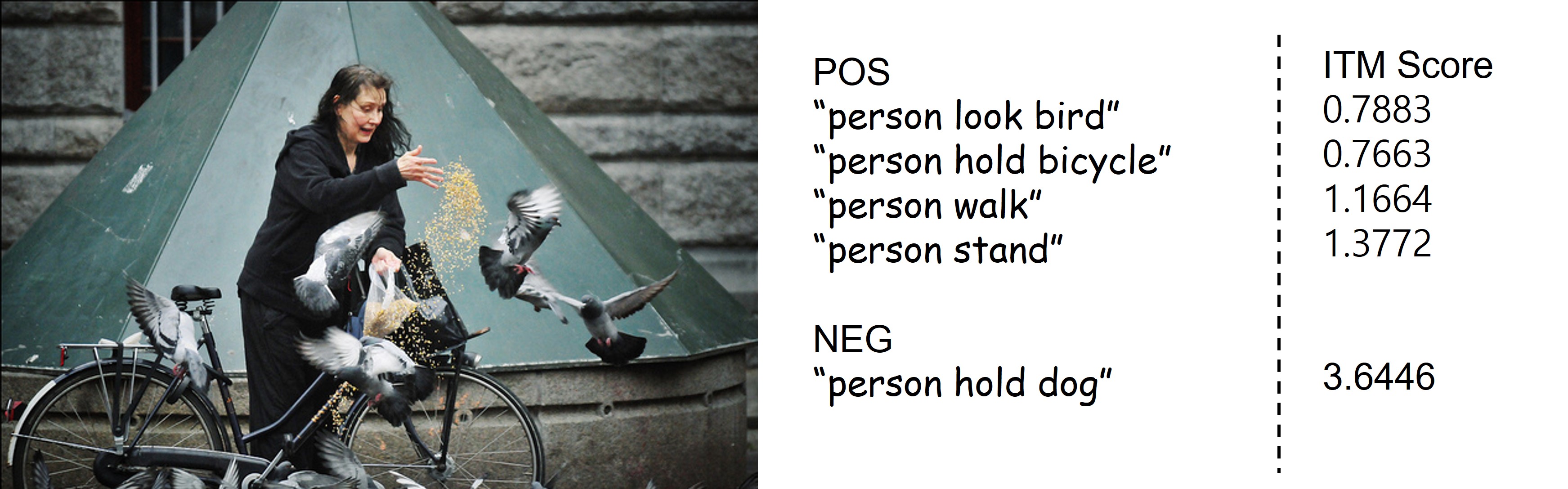}
         \label{fig:three sin x}
    \end{subfigure}
    \caption{Analysis of Image-Text Matching Scores on HOI Detection Benchmarks. This figure visualizes the image-text similarity scores computed between visual input and corresponding grounded sentence prompts.}
    \label{fig:itm}
\end{figure*}

\subsection{Computational Resource}

\begin{table}
\setlength{\tabcolsep}{20pt}
    \centering
    \begin{tabular}{llc}
    \toprule
    \multirow{2}{*}{\text{Method}} & \multirow{2}{*}{\text{Parameters}} & training time  \\
    \cmidrule(lr){3-3}
    & & inference time\\
    \midrule
    \multirow{2}{*}{\text{Baseline\cite{muren}}}  & 41M(DETR) & 21.2 min / epoch \\
    \cmidrule(lr){3-3}
    &  28M(Decoder) & 0.03 sec / frame\\
    \midrule
    \multirow{2}{*}{VLM-HOI}  & \text{430M(DETR+BLIP)} & 50.3 min / epoch \\
    \cmidrule(lr){3-3}
    & \textbf{28M(Learnable)} & 0.03 sec / frame \\
    \bottomrule
    \end{tabular}
    \caption{Comparison of the model size, training time per epoch, and inference time per frame between our proposed method and relevant baseline approaches. Our proposed VLM-HOI incorporates a 361M parameter BLIP, of which only the 28M decoder is learned.}
    \label{tab:computational_resource}
\end{table}

A core component of our approach is the large pre-trained VLM which increases the computational requirements.
As shown in Table~\ref{tab:computational_resource}, we use BLIP with 361M parameters due to hardware constraints, compared to 69M parameters for the baseline HOI network.

The additional VLM computations result in longer training times compared to the baseline - around 2.4$\times$ in our experiments.
However, we found that our method requires fewer training epochs to converge, likely due to the beneficial regularization and optimized initialization from the VLM distillation.

At inference time, our model has the same efficiency as the baseline since the VLM is only used during training for knowledge transfer. 
We only need to perform a single forward pass through the learned HOI network.

\section{Conclusion}

In this paper, we present a novel approach that leverages the capabilities of the Large Vision Language Model (VLM) to enhance Human-Object Interaction (HOI) detection. By utilizing VLM as an objective function in the context of HOI detection, we have successfully quantified the similarity of predicted HOI triplets through Image-Text matching, harnessing VLM's comprehensive understanding of both visual and linguistic modalities. Our experiments have yielded state-of-the-art results in HOI detection accuracy on benchmark datasets, marking a significant advancement in the field. This novel integration of VLM into HOI detection not only showcases the potential of language comprehension capabilities in bridging modalities but also takes a promising stride towards more advanced and interpretable human-object interaction analysis. We hope the findings presented in this paper offer valuable insights and open new avenues for research at the intersection of vision and language.

\section*{Acknowledgments}
This work was supported partly by the Institute of Information \& Communications
Technology Planning \& Evaluation (IITP) grant funded by Korea
government (MSIT) [2021-0-01341, Artificial Intelligent Graduate School
Program (Chung-Ang University)], and partly by Field-oriented Technology Development Project for Customs Administration through National Research Foundation of Korea (NRF) funded by the Ministry of Science \& ICT and Korea Customs Service (2021M3I1A1097911).

\bibliographystyle{splncs04}
\bibliography{main}
\end{document}